\documentclass{article}

\usepackage{times}
\usepackage{graphicx} 
\usepackage{subfigure}
\usepackage{booktabs}

\usepackage{enumitem} 
\setlist{itemsep=.4ex, parsep=.1ex, topsep=.1ex, leftmargin=3ex,
	font=\bfseries}

\usepackage{natbib}

\usepackage{algorithm}
\usepackage{algorithmic}

\usepackage{hyperref}

\usepackage[accepted]{icml2016}

\usepackage{amsfonts}
\usepackage{amssymb}
\usepackage{amsmath}
\usepackage{stmaryrd}
\usepackage{amsthm}

\usepackage[T1]{fontenc}

\def\E{{\mathbb E}}

\def\EE{{\mathbf E}}

\def\RR{{\mathbb{R}}}
\def\Real{{\mathbb{R}}}

\def\x{{\mathbf x}}
\def\f{{\mathbf f}}

\def\c{{\mathbf c}}

\def\v{{\mathbf v}}
\def\X{{\mathbf X}}
\def\I{{\mathbf I}}

\def\A{{\mathbf A}}
\def\B{{\mathbf B}}
\def\b{{\mathbf b}}
\def\C{{\mathbf C}}
\def\F{{\mathbf F}}
\def\D{{\mathbf D}}

\def\d{{\mathbf d}}
\def\b{{\mathbf b}}

\def\x{{\mathbf x}}

\def\Real{{\mathbb R}}

\def\M{{\mathbf M}}

\def\trace{{\mathrm{Tr}\;}}

\floatname{algorithm}{Procedure}
\renewcommand{\algorithmicrequire}{\textbf{Input:}}
\renewcommand{\algorithmicensure}{\textbf{Output:}}

\newcommand{\transpose}{^\top}
\newcommand{\argmin}{\operatornamewithlimits{\mathrm{argmin}}}
\newcommand{\balpha}{\boldsymbol{\alpha}}

\newenvironment{myfont}{\fontfamily{fvs}\selectfont}{\par}

\icmltitlerunning{Dictionary Learning for Massive Matrix Factorization}

\begin{document}

\twocolumn[
\icmltitle{Dictionary Learning for Massive Matrix Factorization}

\icmlauthor{Arthur Mensch}{arthur.mensch@m4x.org}
\icmladdress{Parietal team, Inria, CEA,
 Paris-Saclay University. Neurospin,
 Gif-sur-Yvette, France}

 \icmlauthor{Julien Mairal}{julien.mairal@inria.fr}
 \icmladdress{Thoth team, Inria, Grenoble, France}

 \icmlauthor{Bertrand Thirion}{betrand.thirion@inria.fr}
 \icmlauthor{Ga\"el Varoquaux}{gael.varoquaux@inria.fr}

\icmladdress{Parietal team, Inria, CEA,
 Paris-Saclay University. Neurospin,
 Gif-sur-Yvette, France}

\icmlkeywords{Matrix Factorization, Dictionary Learning, Online, Sparse, Stochastic, Neuroscience, Recommender systems}

\vskip 0.3in
]

\begin{abstract}
Sparse matrix factorization is a popular tool to
obtain interpretable data decompositions, which are also effective to
perform data completion or denoising. Its applicability to large datasets has
been addressed with
online and randomized methods, that reduce the
complexity in one of the matrix dimension, but not in both of
them. In this paper, we tackle very large matrices in both dimensions. We propose a new factorization method that scales gracefully to terabyte-scale datasets. Those
could not be processed by previous algorithms in a reasonable amount of time.
We demonstrate the efficiency of our approach on massive functional Magnetic Resonance
Imaging (fMRI) data, and on matrix completion problems for recommender systems, where we obtain significant speed-ups compared to state-of-the art coordinate descent methods.
\end{abstract}

Matrix factorization is a flexible tool for uncovering latent factors in
low-rank or sparse models.
For instance, building on low-rank structure, it has proven very powerful
for matrix completion, \emph{e.g.}~in recommender systems
\citep{srebro_maximum-margin_2004,candes_exact_2009}. In signal
processing and computer vision, matrix factorization with a sparse
regularization is often called dictionary learning
and has proven very effective for
denoising and visual feature encoding~\citep[see][for a review]{mairal_sparse_2014}.
It is also flexible enough to
accommodate a large set of constraints and regularizations, and has gained
significant attention in scientific domains where interpretability is a key
aspect, such as genetics and
neuroscience~\citep{varoquaux_multi-subject_2011}.

As a widely-used model, the literature of matrix factorization is very
rich and two main classes of formulations have emerged. The first one addresses
an optimization problem involving a convex penalty, such as the trace or max
norms~\cite{srebro_maximum-margin_2004}. These penalties
promote low-rank structures, have strong theoretical guarantees~\citep{candes_exact_2009}, but they
do not encourage sparse factors and lack scalability for very-large
datasets. For these reasons, our paper is focused on a second type of
approach, that relies on nonconvex optimization. Specifically, the motivation of
our work originally came from the need to analyze huge-scale fMRI datasets, and
the difficulty of current algorithms to process them.

To gain scalability, stochastic (or online) optimization methods have been
developed; unlike classical alternate minimization procedures, they learn
matrix decompositions by observing a single matrix column (or row) at each
iteration. In other words, they stream data along one matrix dimension.  Their
cost per iteration is significantly reduced, leading to faster convergence in
various practical contexts.  More precisely, two approaches have been
particularly successful: stochastic gradient descent
\citep[see][]{bottou_large-scale_2010} has been widely used in recommender
systems \citep[see][and references therein]{bell_lessons_2007,
rendle_online-updating_2008,rendle_factorization_2010,blondel_convex_2015}, and
stochastic majorization-minimization methods for dictionary learning with
sparse and/or structured regularization
\cite{mairal_online_2010,mairal_stochastic_2013}.
Yet, stochastic algorithms for dictionary learning are currently unable to deal
efficiently with matrices that are large in both dimensions.



In a somehow orthogonal way, the growth of dataset size has proven to be
manageable by \textit{randomized} methods, that exploit random projections
\citep{johnson_extensions_1984,bingham_random_2001} to reduce data dimension without deteriorating
signal content. Due to the way they are generated, large-scale datasets generally have an intrinsic dimension that is significantly smaller than their ambient
dimension.
Biological datasets \cite{mckeown_analysis_1998} and physical acquisitions with
an underlying sparse structure enabling compressed sensing
\cite{candes_near-optimal_2006} are good examples. In this context, matrix
factorization can be performed by using random summaries of coefficients. Recently, those have been used to compute PCA \citep{halko_finding_2009}, a
classical matrix decomposition technique. Yet, using random projections as a
pre-processing step is not appealing in our applicative context since the factors
learned on reduced data loses interpretability.

\paragraph{Main contribution.}
In this paper, we propose
a dictionary learning algorithm that (i) scales
both in the signal dimension (number of rows) and number of signals (number of
columns), (ii) deals with various structured sparse regularization
penalties, (iii) handles missing values, and (iv) provides an explicit dictionary with easy interpretation.
As such, it is non-trivial extension of the online dictionary
learning method of~\citet{mairal_online_2010}, where, at every iteration, signals are
partially observed with a random mask, and with
low-complexity update rules that depend on the (small) mask size instead of the
signal size.

To the best of our knowledge, our algorithm is the first that enjoys all
aforementioned features; in particular, we are not aware of any other
dictionary learning algorithm that is scalable in both matrix dimensions. For
instance, \citet{pourkamali-anaraki_efficient_2015} use random projection with
k-SVD, a \textit{batch} dictionary learning
algorithm \citep{aharon_k-svd:_2006} that does not scale well in the number of training signals.
Online matrix decomposition in the context of missing values was
also proposed by \citet{szabo_online_2011}, but without scalability in the
signal (row) size.

On a massive fMRI dataset (2TB, $n=2.4\cdot 10^6$, $p=2\cdot 10^5$), we were able to learn interpretable dictionaries in about 10 hours on a single workstation,
an order of magnitude faster than the online
approach of~\citet{mairal_online_2010}.  On collaborative filtering
experiments, where sparsity is not needed, our algorithm
performs favorably well compared to state-of-the-art coordinate descent
methods. In both experiments, benefits for the practitioner were significant.

\section{Background on Dictionary Learning}
{
In this section, we introduce dictionary learning as a matrix
factorization problem, and present stochastic algorithms that observe one
column (or a minibatch) at every iteration. 

\subsection{Problem Statement}
The goal of matrix factorization is to decompose a matrix $\X \in \RR^{p \times
n}$ -- typically $n$ signals of dimension $p$ -- as a product of two
smaller matrices:}
\begin{equation}
    \X \approx \D \A
 \quad \text{with}\quad\D \in \RR^{p \times k}, \;\A \in \RR^{k \times n},
\end{equation}
with potential sparsity or structure requirements on $\D$ and~$\A$. In
statistical signal applications, this is often a dictionary learning
problem, enforcing sparse coefficients $\A$.
In such a case, we call~$\D$ the ``dictionary'' and~$\A$ the sparse codes.
We use this terminology throughout the paper.



Learning the dictionary is typically performed by minimizing a quadratic
data-fitting term, with constraints and/or penalties over the code and the dictionary:
 \begin{equation}
    \min_{\substack{\D \in \mathcal{C} \\ \A=[\balpha_1,\ldots,\balpha_n]
    \in \RR^{k\times n}}} \sum_{i=1}^n
    \frac{1}{2}
    \bigl\|
    \x_i
    - \D \balpha_i
    \bigr\|_2^2 + \lambda \, \Omega(\balpha_i),
 \end{equation}
where $\mathcal{C}$ is
a convex set of $\RR^{p \times k}$, and a $\Omega : \RR^p \rightarrow \RR$ is a
penalty over the code, to enforce structure or sparsity.
In large $n$ and large $p$ settings, typical in recommender systems, this
problem is solved via block coordinate descent, which boils down to
alternating least squares if regularizations on~$\D$ and~$\balpha$ are
quadratic \cite{hastie_matrix_2014}.

\paragraph{Constraints and penalties.} The constraint set $\mathcal{C}$ is traditionally
a technical constraint ensuring that the coefficients~$\balpha$ do not vanish,
making the effect of the penalty~$\Omega$ disappear. However, other constraints can also be
used to enforce sparsity or structure on the dictionary~\citep[see][]{varoquaux_cohort-level_2013}. In our paper,$\mathcal{C}$ is the Cartesian product of a $\ell_1$ or $\ell_2$ norm ball:
\begin{equation}
\mathcal{C} = \{ \D \in \RR^{p \times k}~~~\text{s.t.}~~~ \psi(\d_j) \leq 1
\quad \forall j =1,\ldots,k\},
\label{eq:ball}
\end{equation}
where $\D=[\d_1,\ldots,\d_k]$ and
 $\psi = \Vert \cdot \Vert_1$ or $\psi = \Vert \cdot
\Vert_2$.
The choice of $\psi$ and~$\Omega$ typically offers
some flexibility in the regularization effect that is desired for a specific problem;
for instance, classical dictionary learning uses~$\psi=\| \cdot \|_2$ and~$\Omega=\|\cdot\|_1$,
leading to sparse coefficients~$\balpha$, whereas our experiments on fMRI
uses $\psi=\|\cdot\|_1$ and~$\Omega=\|\cdot\|_2^2$, leading to sparse dictionary
elements~$\d_j$ that can be interpreted as brain activation maps.

\subsection{Streaming Signals with Online Algorithms}

In stochastic optimization, the number of signals~$n$ is assumed to
be large (or potentially infinite), and the dictionary $\D$ can be written as a
solution of
\begin{gather}
\label{eq:expected_loss}
    \min_{\D\in \mathcal{C}} \; f(\D)\quad\text{where}\quad f(\D) = \E_\x \bigl[l(\x, \D)\bigr]
    \\
    \quad l(\x, \D) = \min_{\balpha \in \RR^k} \frac{1}{2} \Vert \x - \D \balpha \Vert_2^2
    + \lambda\, \Omega(\balpha), \notag
\end{gather}
where the signals~$\x$ are assumed to be i.i.d. samples from an unknown probability distribution.
Based on this formulation, \citet{mairal_online_2010} have introduced an online
dictionary learning approach that draws a single signal~$\x_t$ at iteration~$t$ (or a
minibatch), and computes its sparse
code~$\balpha_t$ using the current dictionary~$\D_{t-1}$ according to
\begin{equation}
   \balpha_t \leftarrow \argmin_{\balpha \in \RR^k} \frac{1}{2} \Vert \x_t -\D_{t - 1} \balpha
   \Vert^2_2 + \lambda \,\Omega(\balpha).\label{eq:alpha}
\end{equation}
Then, the dictionary is updated by approximately minimizing the following surrogate
function
\begin{equation}
    \label{eq:original_surrogate}
    g_t (\D) = \frac{1}{t} \sum_{i = 1}^t \frac{1}{2} \bigl\| \x_i - \D \balpha_i
	    \bigr\|_2^2
     + \lambda \,\Omega(\balpha_i),
\end{equation}
which involves the sequence of past signals~$\x_1,\ldots,\x_t$ and the sparse
codes~$\balpha_1,\ldots,\balpha_t$ that were computed in the past iterations
of the algorithm.
The function~$g_t$ is called a ``surrogate'' in the sense that it only
approximates the objective~$f$.
In fact, it is possible to show that it converges to a locally tight upper-bound
of the objective, and that minimizing~$g_t$ at each iteration asymptotically
provides a stationary point of the original optimization problem.
The underlying principle is that of \emph{majorization-minimization}, used
in a stochastic fashion~\citep{mairal_stochastic_2013}.

%
One key to obtain efficient dictionary updates is the observation that
the surrogate~$g_t$ can be summarized by a few sufficient statistics that
are updated at every iteration. In other words, it is possible to
describe~$g_t$ without explicitly storing the past signals~$\x_i$ and
codes~$\balpha_i$ for~$i \leq t$.
Indeed, we may define two matrices $\B_t \in \RR^{p \times
k}$ and $\C_t \in \RR^{k \times k}$
\begin{align}
    \label{eq:original_statistics}
    \C_t = \frac{1}{t} \sum_{i = 1}^t \balpha_i \balpha_i\transpose
    &\qquad \B_t = \frac{1}{t} \sum_{i = 1}^t \x_i \balpha_i\transpose,
\end{align}
and the
surrogate function is then written:
\begin{equation}
    \label{eq:original_surrogate_2}
    g_t (\D) =  \frac{1}{2} \mathrm{Tr} (\D\transpose\D \C_t - \D\transpose \B_t) + \frac{\lambda}{t} \sum_{i = 1}^t \Omega(\balpha_i).
\end{equation}
The gradient of~$g_t$ can be computed as
\begin{equation}
	\nabla_\D g_t(\D)~=~\D \C_t - \B_t.
\end{equation}
Minimization of $g_t$ is performed using block coordinate descent on the columns of~$\D$. In practice,
the following updates are successively performed by cycling over the dictionary elements~$\d_j$ for~$j=1,\ldots,k$
\begin{equation}
   \d_j  \leftarrow   \text{Proj}_{\psi(.) \leq 1} \left[ \d_j - \frac{1}{\C_t[j,j]} \nabla_{\d_j}g_t(\D)\right],\label{eq:proj}
\end{equation}
where $\text{Proj}$ denotes the Euclidean projection over the constraint norm
constraint~$\psi$.  It can be shown that this update corresponds to
minimizing~$g_t$ with respect to~$\d_j$ when fixing the other dictionary
elements~\citep[see][]{mairal_online_2010}.

\subsection{Handling Missing Values} \label{subsec:masked}

Factorization of matrices with missing value have raised a significant interest in
 signal processing and machine learning, especially as a solution
for recommender systems.
In the context of dictionary learning, a similar effort has been made by~\citet{szabo_online_2011} to adapt the framework to missing
values. Formally, a mask $\M$, represented as a binary diagonal matrix in~$\{0,1\}^{p \times p}$, is
associated with every signal $\x$, such that the algorithm can only observe the product
$\M_t\x_t$ at iteration~$t$ instead of a full signal~$\x_t$.
In this setting, we naturally derive the following objective
\begin{gather}
\label{eq:expected_loss_masked}
\min_{\D\in \mathcal{C}} \; f(\D)\quad\text{where}\quad f(\D) = \E_{\x,\M} \bigl[l(\x, \M, \D)\bigr]
    \\
    \quad l(\x, \M, \D) = \min_{\balpha \in \RR^k} \frac{p}{2\trace{\M}}\Vert \M (\x - \D \balpha) \Vert_2^2
    + \lambda \Omega(\balpha), \notag
\end{gather}
where the pairs~$(\x,\M)$ are drawn from the (unknown) data distribution.
Adapting the online algorithm of~\citet{mairal_online_2010} would consist of
drawing a sequence of pairs~$(\x_t,\M_t)$, and building the surrogate
\begin{equation}
   g_t (\D) = \frac{1}{t} \sum_{i = 1}^t  \frac{p}{2s_i}\bigl\| \M_i(\x_i - \D \balpha_i)
	    \bigr\|_2^2
            +  \lambda \,\Omega(\balpha_i),\label{eq:surrogate_masked}
\end{equation}
where~$s_i= \trace{\M_i}$ is the size of the mask and
\begin{equation}
   \balpha_i \in \argmin_{\balpha \in \RR^k} \frac{p}{2s_i}\Vert \M_i (\x_i - \D_{i-1} \balpha) \Vert_2^2
   + \lambda\, \Omega(\balpha).\label{eq:alphaM}
\end{equation}
Unfortunately, this surrogate cannot be summarized by a few sufficient
statistics due to the masks~$\M_i$: some approximations are required.
This is the approach chosen by~\citet{szabo_online_2011}. Nevertheless, the
complexity of their update rules is linear in the \textit{full} signal size~$p$, which makes
them unadapted to the large-$p$ regime that we consider.

\section{Dictionary Learning for Massive Data}

Using the formalism exposed above, we now consider the problem of factorizing a
large matrix~$\X$ in~$\Real^{p \times n}$ into two factors~$\D$ in~$\Real^{p
\times k}$ and~$\A$ in~$\Real^{k \times n}$ with the following setting:
both $n$ and~$p$ are large (greater than~$100\,000$ up to several millions),
whereas $k$ is reasonable (smaller than~$1\,000$ and often near $100$),
which is not the standard dictionary-learning setting; some
entries of~$\X$ may be missing.  Our objective is to recover a good
dictionary~$\D$ taking into account appropriate regularization.

To achieve our goal, we propose to use an objective akin
to~(\ref{eq:expected_loss_masked}), where the masks are now random variables~\textit{independant} from the samples.
In other words, we want to combine ideas of online dictionary learning with
random subsampling, in a principled manner. This leads us to consider an
infinite stream of samples $(\M_t \x_t)_{t \geq 0}$, where the signals~$\x_t$ are i.i.d.
samples from the data distribution --\,that is, a column of~$\X$ selected at
random\,-- and $\M_t$ ``selects'' a random subset of observed entries in~$\X$.
This setting can accommodate missing entries, never selected by the
mask, and only requires loading a subset of $\x_t$ at each iteration.

The main justification for choosing this objective function is that
in the large sample regime $p \gg k$ that we consider,
computing the code~$\balpha_i$ using only a random subset of the data~$\x_t$
according to~(\ref{eq:alphaM}) is a good approximation of the code that may be
computed with the full vector~$\x_t$ in~(\ref{eq:alpha}). This of course
requires choosing a mask that is large enough; in the fMRI dataset, a
subsampling factor of about~$r=10$ --\,that is only~$10\%$ of the entries
of~$\x_t$ are observed\,-- resulted in a similar $10\times$ speed-up (see
experimental section) to achieve the same accuracy as the original approach without subsampling.  This point of view also justifies the
natural scaling factor $\frac{p}{\trace \M}$
introduced in~(\ref{eq:expected_loss_masked}).

An efficient algorithm must address two challenges: (i) performing
dictionary updates that do not depend on~$p$ but only on the mask size;
(ii) finding an approximate surrogate function that can be summarized by a few
sufficient statistics.
We provide a solution to these two issues in the next subsections and present
the method in Algorithm~\ref{alg:dl}.

\subsection{Approximate Surrogate Function}
To approximate the surrogate~(\ref{eq:original_surrogate_2}) from $\balpha_t$ computed in \eqref{eq:alphaM},
we consider $h_t$ defined by
\begin{equation}
    \label{eq:original_surrogate_3}
    h_t (\D) =  \frac{1}{2} \mathrm{Tr} (\D\transpose\D \C_t
    - \D\transpose \B_t) + \frac{\lambda}{t} \sum_{i = 1}^t \frac{s_i}{p} \Omega(\balpha_i)
\end{equation}
with the same matrix~$\C_t$ as in~(\ref{eq:original_surrogate_2}), which is updated as
\begin{equation}
   \C_t \leftarrow \Big(1-\frac{1}{t}\Big)\C_{t-1} + \frac{1}{t} \balpha_t\balpha_t^\top,
\end{equation}
and to replace~$\B_t$ in~(\ref{eq:original_surrogate_2}) by the matrix
\begin{equation}
   \B_t = \Big( \sum_{i=1}^t \M_i \Big)^{-1} \sum_{i=1}^t \M_i\x_i \balpha_i^\top,
\end{equation}
which is the same as~(\ref{eq:original_statistics}) when $\M_i=\I$.
Since~$\M_i$ is a diagonal matrix, $\sum_{i=1}^t \M_i$ is also diagonal and simply ``counts'' how
many times a row has been seen by the algorithm. $\B_t$ thus behaves like $\E_{\x}[\x \balpha(\x, \D_t)^\top]$ for large $t$, as in the fully-observed algorithm. By design, only rows of~$\B_t$ selected by the mask differ from~$\B_{t-1}$. The update can therefore be achieved in $O(s_i k)$ operations:
\begin{equation}
   \B_t = \B_{t-1} + \Big( \sum_{i=1}^t \M_i \Big)^{-1} \left( \M_t\x_t \balpha_t^\top - \M_t \B_{t-1} \right)
\end{equation}
This only requires keeping in memory the diagonal matrix $\sum_{i=1}^t \M_i$,
and updating the rows of~$\B_{t-1}$ selected by the mask.  All operations only
depend on the mask size $s_i$ instead of the signal size $p$.

\subsection{Efficient Dictionary Update Rules}

With a surrogate function in hand, we now describe how to update the
codes~$\balpha$ and the dictionary~$\D$ when only partial access to data is
possible.
The complexity for computing the sparse codes~$\balpha_t$ is obviously
independent from~$p$ since~(\ref{eq:alphaM}) consists in
solving a \textit{reduced} penalized linear regression of $\M_t \x_t$ in~$\Real^{s_t}$ on $\M_t \D_{t-1}$ in~$\Real^{s_t \times k}$.
Thus, we focus here on dictionary update rules.

The naive dictionary update~(\ref{eq:proj2}) has complexity $O(kp)$ due to the
matrix-vector multiplication for computing $\nabla_{\d_j} g_t(\D)$.
Reducing the single iteration complexity of a
factor $\frac{p}{s_t}$ requires reducing the dimensionality of the
dictionary update phase.
We propose two strategies to achieve that, both using block
coordinate descent, by considering
\begin{equation}
   \d_j  \leftarrow   \text{Proj}_{\psi(.) \leq 1} \left[ \d_j - \frac{1}{\C_t[j,j]} \M_t\nabla_{\d_j}h_t(\D)\right],\label{eq:proj2}
\end{equation}
where $\M_t\nabla_{\d_j}h_t(\D)$ is the partial derivative of $h_t$
with respect to the $j$-th column and rows selected by the mask.

\paragraph{Gradient step.}
The update~(\ref{eq:proj2}) represents a classical block coordinate descent step involving particular blocks.
Following~\citet{mairal_online_2010}, we perform one cycle over the columns warm-started on
$\D_{t-1}$.
Formally, the gradient step without projection for the $j$-th component consists of updating the vector~$\d_j$
\begin{equation}
   \begin{split}
      \d_j & \gets \d_j - \frac{1}{\C_t[j,j]} \M_t\nabla_{\d_j}h_t(\D) \\
           & = \d_j - \frac{1}{\C_t[j,j]}(\M_t\D\c_j^t - \M_t\b_j^t),
   \end{split}\label{eq:gradientstep}
\end{equation}
where~$\c_j^t, \b_j^t$ are the~$j$-th  columns of $\C_t$, $\B_t$ respectively.
The update has complexity $O(k s_t)$ since it only involves $s_t$ rows of~$\D$ and only $s_t$ entries of~$\d_j$
have changed.

\paragraph{Projection step.}
Block coordinate descent algorithms require orthogonal
projections onto the constraint set~$\mathcal C$.  In our case, this amounts to
the projection step on the unit ball corresponding to the norm~$\psi$
in~(\ref{eq:proj2}).  The complexity of such a projection is usually $O(p)$
both for~$\ell_2$ and~$\ell_1$-norms \citep[see][]{duchi_efficient_2008}.
We consider here two strategies.

\subparagraph{Exact lazy projection for~$\ell_2$.}
When~$\psi=\ell_2$, it is possible to perform the projection implicitly with
complexity $O(s_t)$.  The computational trick is to notice that the projection
amounts to a simple rescaling operation
%
\begin{align}
	\label{eq:l2_proj}
	\d_j \gets \frac{\d_j}{\max(1, \Vert \d_j \Vert_2)},
\end{align}
which may have low complexity if the dictionary elements~$\d_j$ are stored in
memory as a product $\d_j{=}\f_j/\max(1,l_j)$ where $\f_j$ is in~$\Real^{p}$
and~$l_j$ is a rescaling coefficient such that~$l_j = \|\f_j\|_2$. We code the gradient
 step~(\ref{eq:gradientstep}) followed by $\ell_2$-ball projection by the updates
\begin{equation}
	\begin{split}
      n_j &\leftarrow  \|\M_j\f_j\|_2^2\\
	  \f_j &\leftarrow \f_j - \frac{\max(1,l_j)}{\C_t[j,j]}(\M_t\D\c_j^t - \M_t\b_j^t) \\
      l_j & \leftarrow \sqrt{ l_j^2 - n_j + \|\M_j\f_j\|_2^2}
  \end{split}
\end{equation}
Note that the update of~$\f_j$ corresponds to the gradient step without
projection~(\ref{eq:gradientstep}) which costs~$O(k s_t)$, whereas the norm
of~$\f_j$ is updated in $O(s_t)$ operations.
The computational complexity is thus independent of~$p$ and the only price to pay
is to rescale the dictionary elements on the fly, each time we need access to them.

\begin{algorithm}[t]
\begin{algorithmic}
    \REQUIRE \text{Initial dictionary:} $\D_0 \in \RR^{p \times k}$,
    \;\text{tolerance:} $\epsilon$
    \STATE $\C_0 \gets 0 \in \RR^{k \times k}$; \qquad $\B_0 \gets 0 \in \RR^{p \times k}$;
    \STATE $\EE_0 \gets 0 \in \RR^{p\times p }$ (diagonal); \qquad $t \gets 1$;
    \REPEAT
    \STATE Draw a pair $(\x_t,\M_t)$;
    \STATE $\balpha_t \!\! \gets \!\! \argmin_{\balpha}\frac{1}{2}
        \Vert \M_t (\x_t \!-\! \D_{t-1} \balpha) \Vert_2^2 \!+\! \lambda \frac{\trace \M_t}{p} \Omega
        (\balpha)$; \\

        \STATE $\EE_t \gets \EE_t + \M_t$;

        \STATE $\A_t \gets (1 - \frac{1}{t}) \A_{t-1} + \frac{1}{t}
        \balpha_t {\balpha_t}\transpose$;

        \STATE $\B_t \gets \B_{t-1} + \EE_t^{-1} (
        \M_t \x_t {\balpha_t}\transpose - \M_t \B_{t-1})$;

        \STATE $\D_t \gets \mathtt{dictionary\_update}(\B_t,\C_t, \D_{t-1}, \M_t)$;

    \UNTIL{$| \frac{h_{t-1}(\D_{t-1})}{h_t(\D_t)} -1 | < \epsilon$}

    \ENSURE $\D$
\end{algorithmic}
\caption{Dictionary Learning for Massive Data}
\label{alg:dl}
\end{algorithm}

\subparagraph{Exact lazy projection for~$\ell_1$.}
The case of~$\ell_1$ is slightly different but can be handled in a similar
manner, by storing an additional scalar~$l_j$ for each dictionary element~$\d_j$.
More precisely, we store a vector~$\f_j$ in~$\Real^p$ such that $\d_j~=~\text{Proj}_{\psi(.) \leq 1}[ \f_j]$,
and a classical result~\citep[see][]{duchi_efficient_2008} states that there
exists a scalar~$l_j$ such that
\begin{equation}
   \d_j = S_{l_j}[\f_j], \quad S_\lambda(u) = \text{sign}(u).
\max(|u|-\lambda,0)
\label{eq:softthreshold}
\end{equation}
where~$S_\lambda$ is the soft-thresholding operator, applied elementwise to the entries of~$\f_j$.
Similar to the case~$\ell_2$, the ``lazy'' projection consists of tracking the
coefficient~$l_j$ for each dictionary element and updating it after each
gradient step, which only involves $s_t$ coefficients.
For such sparse updates followed by a projection onto the~$\ell_1$-ball,
\citet{duchi_efficient_2008} proposed an algorithm to find the
threshold~$l_j$ in $O(s_t \log(p))$ operations.
The lazy algorithm involves using particular data structures such as red-black trees
and is not easy to implement; this motivated us to investigate another
simple heuristic that also performs well in practice.

\subparagraph{Approximate low-dimension projection.}
The heuristic consists in performing the projection by forcing the coefficients outside the mask not to change. This results in the orthogonal projection of each $\d_j$ on
${\mathcal T}_{t,j} = \{\d\ ~~\text{s.t.}~~\psi(\d)\leq~1,\,(\I - \M_t)\d = (\I - \M_t)\d_j^{t-1}\}$, which is a subset of the original constraint set $\psi(\cdot) \leq 1$.
%

\begin{algorithm}[t]
\begin{algorithmic}
    \REQUIRE $\B, \C, \D, \M$
    \FOR{$j \in 1,\ldots,k$}
    	\STATE $\d_j \gets \d_j  - \frac{1}{\C[j, j]}(\M\D\c_j - \M\b_j)$;
		\IF{approximate projection}
                \STATE $\v_j \gets \text{Proj}_{{\mathcal T}_j} \left[\M\d_j \right]$,\\
                (see main text for the definition of ${\mathcal T}_j$);
                \STATE $\d_j \gets \d_j + \M\v_j - \M\d_j$;
		\ELSIF{exact (lazy) projection}
                \STATE or $\d_j \gets \text{Proj}_{\psi(.) \leq 1} \left[\d_j \right]$;
		\ENDIF
    \ENDFOR
\end{algorithmic}
\caption{Dictionary Update}
\label{alg:dict_partial}
\end{algorithm}

All the computations require only 4 matrices kept in memory
$\B$, $\C$, $\D$, $\EE$ with additional $\F$, $\mathbf l$ matrices and vectors for the exact
projection case, as summarized in Alg. \ref{alg:dl}.

\subsection{Discussion}
\label{sec:discussion}

\begin{figure*}[t]
  \centering
  \includegraphics{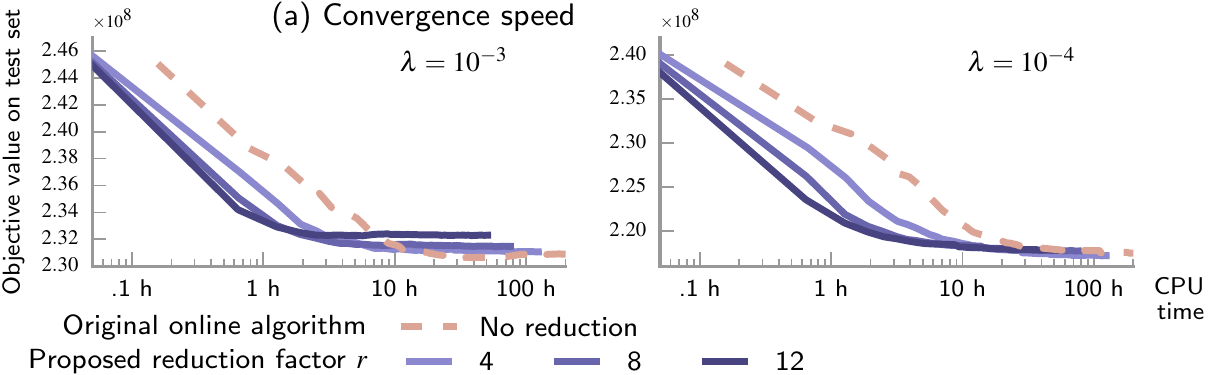}%
  \includegraphics{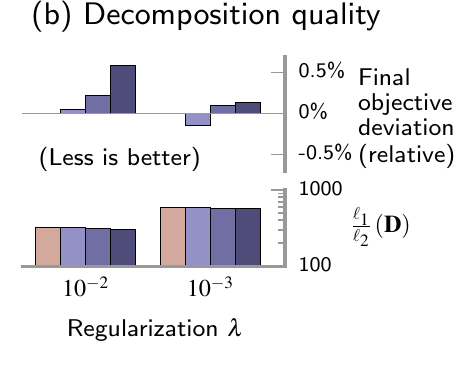}
  \caption{ \textbf{Acceleration of sparse matrix factorization} with random subsampling on the HCP dataset (\textbf{2TB}). Reducing streamed data with stochastic masks permits $10\times$ speed-ups without deteriorating goodness of fit on test data nor alterating sparsity of final dictionary.} 
   \label{fig:hcp_curve}
   \vspace{-.8em}
\end{figure*}

\paragraph{Relation to classical matrix completion formulation.}
Our model
is related to the classical $\ell_2$-penalized matrix completion model \citep[\textit{e.g.}][]{bell_lessons_2007} we rewrite
\begin{equation}
   \sum_{i=1}^n\Vert \M_i (\x_i -\D\transpose\balpha_i\!)
\Vert_2^2 + \lambda s_i \Vert \balpha_i \Vert_2^2 + \lambda \Vert
(\sum_{i=1}^n \M_i)^{\frac{1}{2}} \D \Vert_2^2
	\label{eq:mmmf}
\end{equation}
With quadratic regularization on~$\D$ and~$\A$ --\,that is, using
$\Omega=\|.\|_2^2$ and~$\psi=\|.\|_2$\,-- \eqref{eq:expected_loss_masked} only
differs in that it uses a penalization on $\D$ instead of a constraint. \citet{srebro_maximum-margin_2004}
 introduced the trace-norm regularization to solve a convex problem equivalent to \eqref{eq:mmmf}.
The major difference is that we adopt a non-convex optimization strategy, thus
losing the benefits of convexity, but gaining on the other hand the possibility
of using stochastic optimization.

\paragraph{Practical considerations.} Our algorithm can be slightly modified to use
weights $w_t$ that differ from $\frac{1}{t}$ for $\B$ and $\C$, as advocated by \citet{mairal_stochastic_2013}.
It also proves beneficial to perform code computation on mini-batches of masked
samples. Update of the dictionary is performed on the rows that are seen at
least once in the masks $(\M_t)_{\textrm{batch}}$.

\section{Experiments}

The proposed algorithm was designed to handle massive datasets:
masking data enables streaming a sequence $(\M_t \x_t)_t$ instead of
$(\x_t)_t$, reducing single-iteration computational complexity and IO stress of a factor $r = \frac{p}{\E(\trace \M)}$,
while accessing an accurate description of the data. Hence, we analyze in
detail how our algorithm improves performance for
sparse decomposition of fMRI datasets. Moreover, as it relies on data
masks, our algorithm is well suited for matrix completion,
to reconstruct a data stream $(\x_t)_t$ from the masked stream
$(\M_t \x_t)_t$.
We demonstrate the accuracy of our algorithm on explicit
recommender systems and show considerable computational speed-ups
compared to an efficient coordinate-descent based algorithm.

We use \textit{scikit-learn} \cite{pedregosa_scikit-learn:_2011} in experiments,
and have released a python package\footnote{\url{http://github.com/arthurmensch/modl}} for reproducibility.

\subsection{Sparse Matrix Factorization for fMRI}

\paragraph{Context.}Matrix factorization has long been used on
functional Magnetic Resonance Imaging \cite{mckeown_analysis_1998}. Data are
temporal series of 3D images of brain activity, to decompose in spatial modes
capturing regions that activate together. The matrices to decompose are dense
and heavily redundant, both spatially and temporally: close voxels and
successive records are correlated. Data can be huge:
we use the whole HCP dataset \cite{van_essen_wu-minn_2013}, with $n=2.4\cdot 10^6$ (2000 records, 1\,200 time points) and $p=2\cdot10^5$, totaling 2\,TB of dense
data.

Interesting dictionaries for neuroimaging capture spatially-localized
components, with a few brain regions. This can be obtained by enforcing
sparsity on the dictionary: in our formalism, this is achieved with
$\ell_1$-ball
projection for $\D$. We set $\mathcal{C} = \mathcal{B}_1^k$, and $\Omega = \Vert
\cdot\Vert_2^2$. Historically, such decomposition have been obtained
with the classical
dictionary learning objective on \textit{transposed} data
\cite{varoquaux_cohort-level_2013}: the code $\A$ holds sparse spatial
maps and voxel time-series are streamed. However, given the size of $n$ for our
dataset, this method is not usable in practice.

\begin{figure*}
\begin{myfont}
\hspace*{-.028\linewidth}%
\includegraphics[width=.21\linewidth]{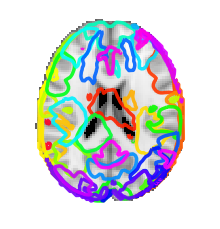}%
\llap{\raisebox{.004\linewidth}{\rlap{\footnotesize\sffamily {\bfseries
235\,h}
run time}\hspace*{.145\linewidth}}}%
\hspace*{-.08\linewidth}%
\includegraphics[width=.22\linewidth]{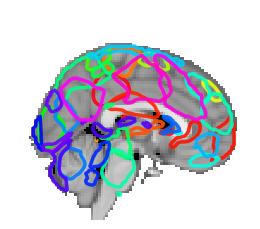}%
\llap{\raisebox{.193\linewidth}{\rlap{\scriptsize\sffamily\scalebox{1.1}{%
Original online algorithm}}\hspace*{.195\linewidth}}}%
\llap{\raisebox{.17\linewidth}{\rlap{\scriptsize\sffamily\scalebox{1.1}{%
1 full epoch}}\hspace*{.17\linewidth}}}%
\hspace*{-.02\linewidth}%
\hfill%
\includegraphics[width=.21\linewidth]{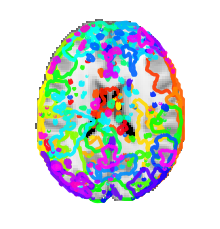}%
\llap{\raisebox{.004\linewidth}{\rlap{\footnotesize\sffamily {\bfseries
10\,h}
run time}\hspace*{.145\linewidth}}}%
\hspace*{-.08\linewidth}%
\includegraphics[width=.22\linewidth]{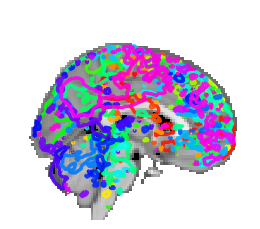}%
\llap{\raisebox{.193\linewidth}{\rlap{\scriptsize\sffamily\scalebox{1.1}{%
Original online algorithm}}\hspace*{.195\linewidth}}}%
\llap{\raisebox{.17\linewidth}{\rlap{\scriptsize\sffamily\scalebox{1.1}{%
$\frac{1}{24}$ epoch}}\hspace*{.17\linewidth}}}%
\hspace*{-.02\linewidth}%
\hfill%
\includegraphics[width=.21\linewidth]{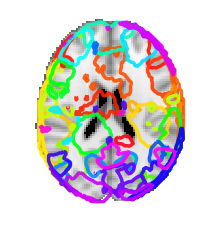}%
\llap{\raisebox{.004\linewidth}{\rlap{\footnotesize\sffamily {\bfseries
10\,h}
run time}\hspace*{.145\linewidth}}}%
\hspace*{-.08\linewidth}%
\includegraphics[width=.22\linewidth]{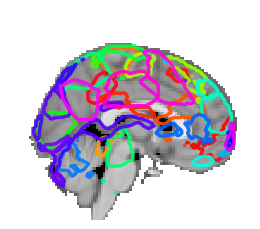}%
\llap{\raisebox{.193\linewidth}{\rlap{\scriptsize\sffamily\scalebox{1.1}{%
{\bfseries Proposed algorithm}}}\hspace*{.195\linewidth}}}%
\llap{\raisebox{.17\linewidth}{\rlap{\scriptsize\sffamily\scalebox{1.1}{%
$\frac{1}{2}$ epoch, reduction $r{=}12$}}\hspace*{.184\linewidth}}}%
\hspace*{-.01\linewidth}%
\vspace*{-.01\linewidth}%
\end{myfont}

\caption{\textbf{Brain atlases}: outlines of each map at half the maximum value ($\lambda = 10^{-4}$).
\textbf{Left}: the reference algorithm on the full dataset.
\textbf{Middle}: the reference algorithm on a twentieth of the dataset.
\textbf{Right}: the proposed algorithm with a similar
run time: half the dataset and $r = 9$.
Compared to a full run of the baseline algorithm,
the figure explore two possible strategies to decrease computation time:
processing less data (middle), or our approach (right). Our approach
achieves a result closer to the gold standard in a given time budget.
\label{fig:brains}}
\end{figure*}

Handling such volume of data sets new constraints. 
First,
efficient disk access becomes critical for speed. In our case,
learning the dictionary is done by accessing the data in row batches, which is
coherent with fMRI data storage: no time is lost seeking data on
disk. Second, reducing IO load on the storage is also
crucial, as it lifts bottlenecks that appear when many processes
access the same storage at the same time, \emph{e.g.} during
cross-validation on $\lambda$ within a supervised pipeline. Our approach
reduces disk usage by a factor $r$. Finally,
parallel methods based on message passing, such as
asynchronous coordinate descent, are unlikely to be efficient given the network /
disk bandwidth that each process requires to load data. This makes it crucial to design
efficient sequential algorithms.
%

\paragraph{Experiment}We quantify the effect of random subsampling for sparse
matrix factorization, in term of speed and accuracy.
A natural performance evaluation is to measure an
empirical estimate of the loss $l$ defined in Eq.~\ref{eq:expected_loss} from \textit{unseen} data, to
rule out any overfitting effect. For this, we evaluate $l$ on a test set $(\x_i)_{i < N}$.
%
%
Pratically, we sample $(\x_t)_t$ in a pseudo-random manner: we randomly select a
record, from where we select a random batch of rows $\x_t$ -- we use a batch
size of $40$, empirically found to be efficient. We load $\M_t \x_t$ in memory
and perform an iteration of the algorithm. The mask sequence is sampled by breaking random permutation vectors into chunks of size $p / r$.

\begin{figure}
\includegraphics{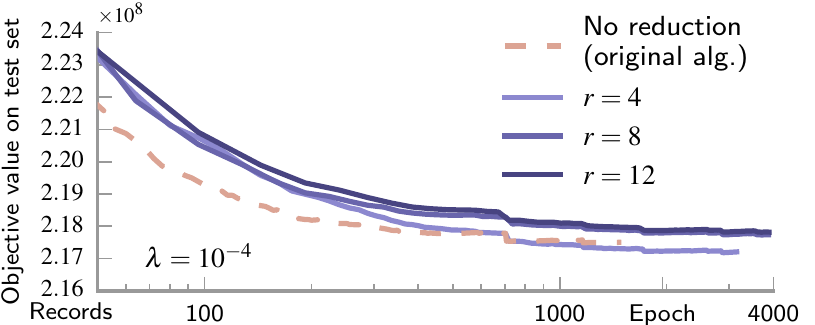}
\vspace{-1.5em}
\caption{\textbf{Evolution of objective function with epochs} for three reduction factors. Learning speed per epoch is little reduced
 by stochastic subsampling, despite the speed-up factor it provides.
\label{fig:epoch}}
\end{figure}

\paragraph{Results}Fig.~\ref{fig:hcp_curve}(a) compares our algorithm with
subsampling ratios $r$ in $\{4, 8, 12\}$ to vanilla online dictionary learning
algorithm $(r = 1)$, plotting trajectories of the \emph{test} objective against real CPU
time. There is no obvious choice of~$\lambda$ due to the unsupervised nature of
the problem: we use $10^{-3}$ and $10^{-4}$, that bounds the range of $\lambda$ providing
interpretable dictionaries.

First, we observe the convergence of the objective
function for all tested $r$, providing evidence that the approximations made in the derivation of update rules does not break convergence for such $r$.
Fig. \ref{fig:hcp_curve}(b) shows the validity of the
obtained dictionary relative to the reference output: both objective function
and $\ell_1{/}\ell_2$ ratio -- the relevant value to measure sparsity in our setting -- are comparable to the baseline values, up to $r = 8$. For high regularization and $r = 12$, our algorithm
 tends to yield somewhat sparser solutions
 ($5\%$ lower $\ell_1{/}\ell_2$) than the original algorithm, due to the approximate $\ell_1$-projection we perform. Obtained maps still proves as interpretable as with baseline algorithm.

Our algorithm proves much faster than the original
one in finding a good dictionary. Single iteration time is indeed reduced by a
factor $r$, which enables our algorithm to go over a single epoch $r$~times~faster
than the vanilla algorithm and capture the variability of
the dataset earlier. To quantify speed-ups, we plot the empirical
objective value of $\D$ against the number of observed records in Fig.~\ref{fig:epoch}. For $r \leq 12$, increasing $r$ little reduces convergence
speed per epoch: random subsampling does not shrink much the quantity of information learned at each iteration.

This brings a near $\times r$ speed-up factor: for high and low regularization respectively, our algorithm converges in $3$ and $10$ hours with subsampling factor $r = 12$, whereas
the vanilla online algorithm requires about $30$ and $100$ hours.
Qualitatively, Fig.~\ref{fig:brains} shows that with the same time
budget, the proposed reduction approach with $r=12$ on half of the data
gives better results than processing a small fraction of the data without
reduction: segmented regions are less noisy and closer to processing the full data.

These results advocates the use of a subsampling rate of $r\approx10$ in this
setting. When sparse matrix decomposition is part of a supervised pipeline with
scoring capabilities, it is possible to find $r$ efficiently: start by setting
it dereasonably high and decrease it geometrically until supervised performance
(\textit{e.g.} in classification) ceases to improve.

\subsection{Collaborative Filtering with Missing Data}

\begin{figure*}[ht]
  \centering
  \includegraphics{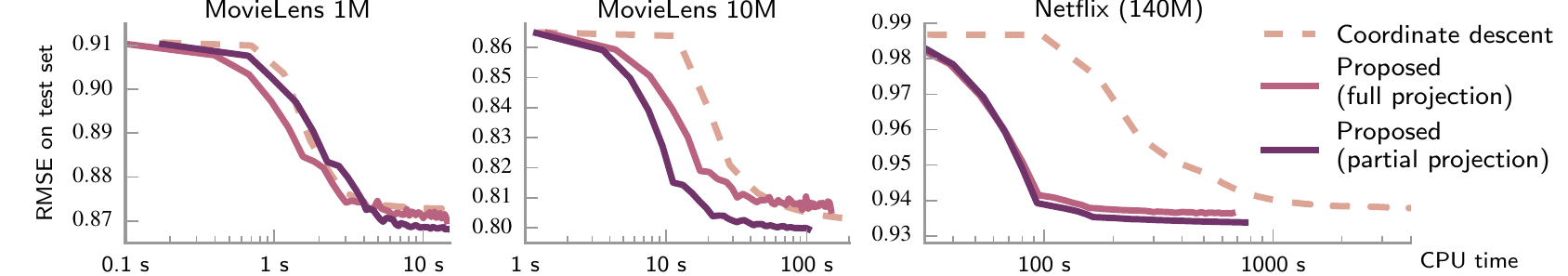}
  \vspace{-2em}
  \caption{\textbf{Learning speed for collaborative filtering}
for datasets of different size: the larger the dataset, the greater
our speed-up.
  \label{fig:bench_rec}}
\end{figure*}

We validate the performance of the proposed algorithm on recommender systems
for explicit feedback, a well-studied matrix completion problem.
We evaluate the scalability of our method
on datasets of different dimension: MovieLens 1M, MovieLens
10M, and 140M ratings Netflix dataset.

We compare our algorithm to a coordinate-descent based method
\cite{yu_scalable_2012}, that provides state-of-the art convergence time
performance on our largest dataset. Although stochastic gradient descent methods for
matrix factorization can provide slightly better single-run performance \citep{takacs_scalable_2009}, these are notoriously hard
to tune and require a precise grid search to uncover a working
schedule of learning rates. In contrast, coordinate descent methods do not
require any hyper-parameter setting and are therefore more efficient in
practice.
We benchmarked various recommender-system codes
(\textit{MyMediaLite}, \textit{LibFM}, \textit{SoftImpute},
\textit{spira}\footnote{https://github.com/mblondel/spira}), and chose
coordinate descent algorithm from \textit{spira} as it was by far the fastest.

\paragraph{Completion from dictionary $\D_t$.}
We stream user ratings to our algorithm: $p$ is the number of movies and $n$ is
the number of users. As $n \gg p$ on Netflix dataset, this increases
the benefit of using an online method. We have observed comparable prediction
performance streaming item ratings.
Past the first epoch, at iteration $t$, every column $i$ of
$\X$ can be predicted by the last code $\balpha_{l(i, t)}$ that was computed
from this column at iteration $l(i, t)$. At iteration $t$,
for all $i < [n]$, $\x_i^\textrm{pred} = \D\balpha_{l(i, t)}$. Prediction thus only requires an additional matrix computation
 after the factorization.

\paragraph{Preprocessing.} Successful prediction should take into account user
and item biases. We compute these biases on train data following \citet{hastie_matrix_2014} (alternated debiasing). We use them to center the samples $(\x_t)_t$ that are streamed to
 the algorithm, and to perform final prediction.

%
\paragraph{Tools and experiments.} Both baseline and proposed algorithm are implemented in a
computationally optimal way, enabling fair comparison based on CPU time. Benchmarks were run using a single 2.7\,GHz Xeon CPU, with
 a 30 components dictionary. For Movielens datasets, we use a random
$25\%$ of data for test and the rest for training. We average results
on five train/test split for MovieLens in Table~\ref{table:rmse}. On Netflix,
the probe dataset is used for testing. Regularization parameter
$\lambda$ is set by cross-validation on the training set: the training
data is split 3 times, keeping $33\%$ of Movielens datasets for evaluation and $1\%$ for
Netflix, and grid search is performed on 15 values of $\lambda$ between $10^{-2}$ and $10$. We assess the
quality of obtained decomposition by measuring the root mean square error
(RMSE) between prediction on the test set and ground truth. We use mini-batches of size $\frac{n}{100}$.

\begin{table}
	\small%
	\centering%
	\caption{\textbf{Comparison of performance and convergence time} for online masked matrix factorization and coordinate descent method.
	Convergence time: score is under 0.1\% deviation from final root mean squared error on test set -- 5 runs average. CD: coordinate descent; MODL: masked online dictionary learning.
	\label{table:rmse}}
	\vspace{.1in}
	\setlength{\tabcolsep}{5pt}%
	\begin{tabular}{lcclll}
		\toprule
		Dataset  & \multicolumn{2}{c}{Test RMSE} & \multicolumn{2}{c}{Convergence time} & Speed\\
		\cmidrule(r){2-3}\cmidrule(r){4-5}
		  & CD     & \textbf{MODL}               & CD      & \textbf{MODL} & ~~-up\\
		\midrule
		ML 1M    & $0.872$ & $\mathbf{0.866}$ & $\mathbf{6\,s}$ & $8\,s$ & $\times0.75$\\
		ML 10M   & $0.802$ & $\mathbf{0.799}$ & $223\,s$        & $\mathbf{60\,s}$ &$\times3.7$\\
		NF (140M) & $0.938$ & $\mathbf{0.934}$ & $1714\,s$       & $\mathbf{256\,s}$ & $\times6.8$ \\
		\bottomrule
	\end{tabular}%
	\vspace{-.2em}
\end{table}

\paragraph{Results.}We report the evolution of test RMSE along time in
Fig.~\ref{fig:bench_rec}, along with its value at convergence and numerical convergence
time in Table~\ref{table:rmse}. Benchmarks are performed on the final run,
after selection of parameter $\lambda$.

The two variants of the proposed method
converge toward a solution that is at least as good as that of
coordinate descent, and slightly better on Movielens 10M and Netflix.
Our algorithm brings a substantial performance improvement on medium
and large scale datasets. On Netflix, convergence is almost reached in $4$
minutes (score under $0.1\%$ deviation from final RMSE), which makes our method $6.8$ times faster than coordinate descent.
Moreover, the relative performance of our algorithm increases with
dataset size. Indeed, as datasets grow,
less epochs are needed for our algorithm to reach convergence (Fig.~\ref{fig:bench_rec}). This is a
significant advantage over coordinate descent, that requires a stable number of
cycle on coordinates to reach convergence, regardless of dataset size.
The algorithm with partial projection performs slightly better. This can be explained by the extra regularization on $(\D_t)_t$ brought by this heuristic.

\begin{figure}[hbt]
  \includegraphics{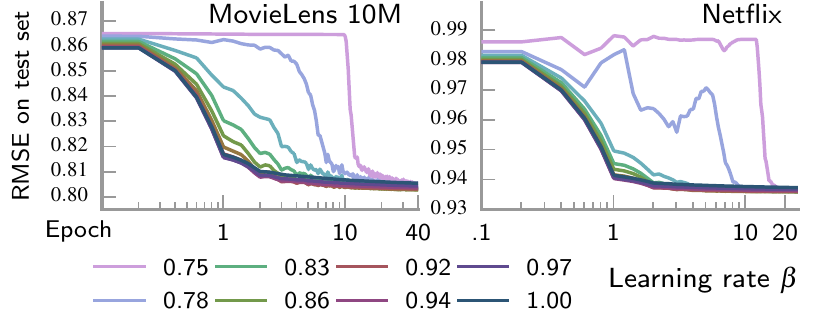}
  \vspace{-.5em}
  \caption{\textbf{Learning weights}:
on two different datasets, optimal convergence is obtained for $\beta \in [.85,\; .95]$, predicted by theory.
  \label{fig:learning_rate}}
  \vspace{-.5em}
\end{figure}

\paragraph{Learning weights.}
Unlike SGD, and similar to the vanilla online dictionary learning algorithm, our method does not
critically suffer from hyper-parameter tuning. We tried weights
$w_t = \frac{1}{t^\beta}$ as described in Sec.~\ref{sec:discussion}, and observed that a range of $\beta$ yields fast convergence.
Theoretically, \citet{mairal_stochastic_2013} shows that stochastic
majorization-minimization converges when $\beta~\in~(.75, 1]$. We verify this empirically, and obtain optimal convergence speed for $\beta \in
[.85, 0.95]$.
(Fig.~\ref{fig:learning_rate}). We report results for $\beta = 0.9$.

\section{Conclusion}

Whether it is sensor data, as fMRI, or e-commerce databases, sample sizes and
number of features are rapidly growing, rendering current matrix
factorization approaches  intractable. We have introduced a online algorithm that leverages
random feature subsampling, giving up to 8-fold speed and memory gains on
large data. 
Datasets are getting bigger,
and they often come with more redundancies. Such approaches
blending online and randomized methods will yield even larger
speed-ups on next-generation data.

\section*{Acknowledgements} The research leading to these results was
supported by the ANR (MACARON project, ANR-14-CE23-0003-01 -- NiConnect
project, ANR-11-BINF-0004NiConnect) and has received funding from the
European Union Seventh Framework Programme (FP7/2007-2013) under grant
agreement no. 604102 (HBP).

\bibliography{bibliography}
\bibliographystyle{icml2016}
\end{document}